\documentclass[runningheads]{llncs}
\usepackage[T1]{fontenc}
\usepackage{graphicx}
\usepackage{algorithm}
\usepackage{algpseudocode}
\usepackage{amsmath}
\usepackage{amssymb}
\usepackage{array}
\usepackage{multirow}
\usepackage{subfigure}
\usepackage{caption}    
\usepackage{subcaption} 
\usepackage{placeins} 
\usepackage{float} 

\begin{document}
\title{Masked Subspace Clustering Methods}
%
%
\author{Jiebo Song\inst{1}\orcidID{0000-0002-6124-2850} \and
Huaming Ling\inst{2}\orcidID{0000-0002-1728-7314} }
\authorrunning{J. Song et al.}
%
\institute{Beijing Institute of Mathematical Sciences and Applications, Huairou, Beijing, China
\email{songjiebo@bimsa.cn}\\
\and
Tsinghua University, Haidian, Beijing, China\\
\email{linghm18@mails.tsinghua.edu.cn}}
\maketitle              
\begin{abstract}
To further utilize the unsupervised features and pairwise information, we propose a general Bilevel Clustering Optimization (BCO) framework to improve the performance of clustering. And then we introduce three special cases on subspace clustering with two different types of masks. At first, we reformulate the original subspace clustering as a Basic Masked Subspace Clustering (BMSC), which reformulate the diagonal constraints to a hard mask. Then, we provide a General Masked Subspace Clustering (GMSC) method to integrate different clustering via a soft mask. Furthermore, based on BCO and GMSC, we induce a learnable soft mask and design a Recursive Masked Subspace Clustering (RMSC) method that can alternately update the affinity matrix and the soft mask. Numerical experiments show that our models obtain significant improvement compared with the baselines on several commonly used datasets, such as MNIST, USPS, ORL, COIL20 and COIL100.

\end{abstract}
\section{Introduction}
Clustering is one of the most fundamental unsupervised learning in machine learning, which aims at grouping similar data points into the same cluster, and distant data points into different clusters~\cite{C1}. Clustering algorithms are widely used in areas without labeled information, especially biological and medical fields. Because of the unlabeled data and high-dimensional data space, it is still a big challenge to design a clustering algorithm. 

There are two typical clustering methods. First, the distance-based method, for example, K-means~\cite{kmeans}. Due to the simplicity and fast speed, Kmeans became one of the most widely used clustering methods. However, K-means always fails in high-dimensional clustering cases since the distance is not enough to character the feature of the data. Second, the relation-based method, such as subspace clustering~\cite{SSC} and graph convolutional network~\cite{GCN}. They use an affinity matrix or an adjacent matrix to present the pairwise relationship between different data points, then make a feature representation based on the global relation, which outperforms in the case of high-dimension data. 

Moreover, spectral clustering~\cite{Spectral} is a hybrid approach that integrates both. It begins by constructing an adjacency matrix using Gaussian distances or existing pairwise constraints. A spectral embedding is then learned by minimizing the smallest singular values, followed by K-means clustering on the embedding to form clusters. While relation-based algorithms are crucial for high-dimensional clustering, their fixed matrices often result in suboptimal performance in complex scenarios. To address this, we propose a novel scheme with a learnable pairwise relation, significantly enhancing clustering performance.

\begin{figure}[t]
\hfill
\subfigure[If the cluster number is greater than 2, it seems easier to distinguish whether two samples belong to the same class than which classes they are.]{\includegraphics[width=4.5cm]{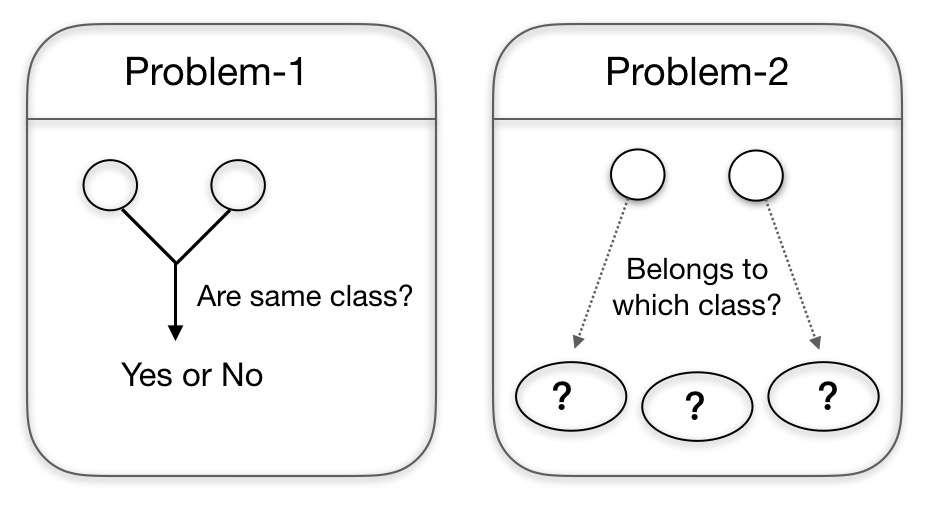}
\label{fig:ca_ba}}
\hfill
\subfigure[(left): a standard diagonal block matrix, where yellow means the pairs have the same class, blue means false. (right): a matrix with a low value of CA but a high value of BCA.]{\includegraphics[width=6cm]{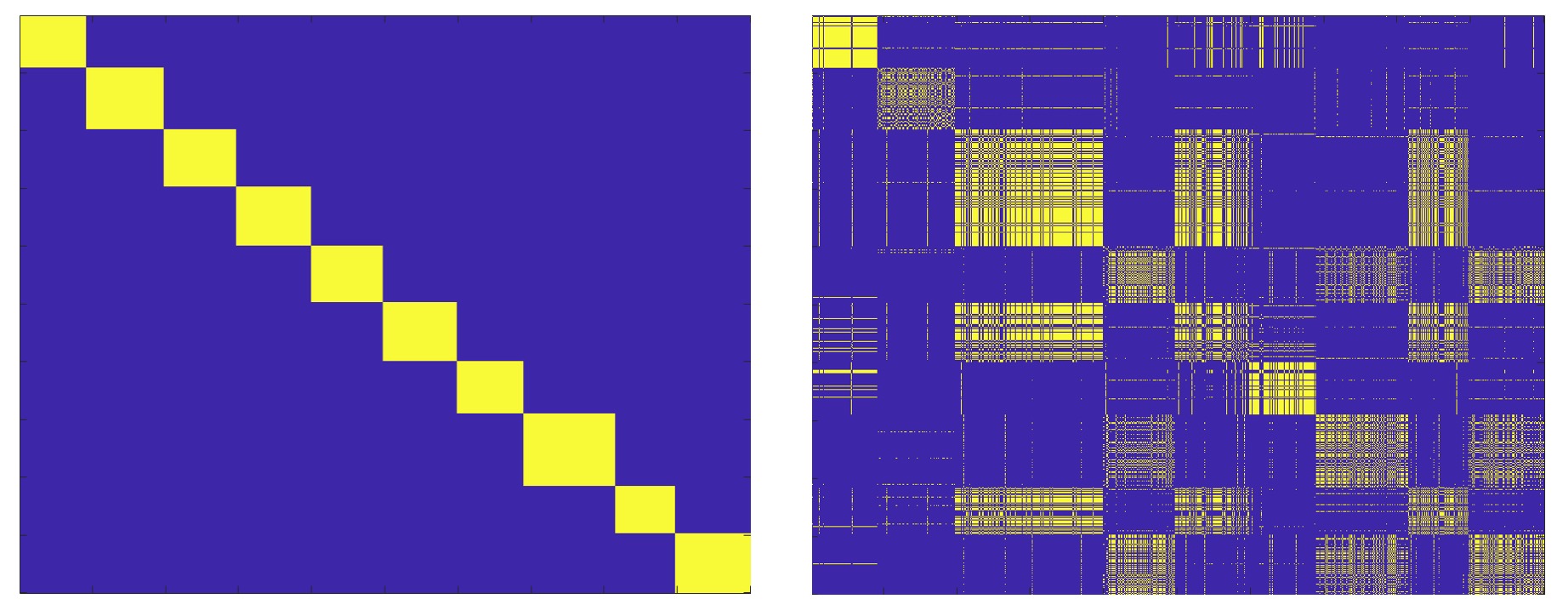}
\label{fig:bce}}
\hfill
\caption{Interesting phenomenons: when the number of clusters is greater than 2, it is easier to determine whether two samples belong to the same cluster than to determine which specific cluster a sample belongs to.}
\end{figure}

We found an interesting phenomenon: as shown in Fig.\ref{fig:ca_ba}, if the cluster number is greater than 2, it seems easier to distinguish whether two samples belong to the same class than which clusters they are. Regarding the \textbf{Clustering Accuracy} as CA, the \textbf{Binary Clustering Accuracy} on whether two samples are the same cluster as BCA. The second matrix in Fig.\ref{fig:bce} shows a clustering method has a lower value of CA (49\%) on 10 classes but with a higher value of BCA(86\%).
Most research focus on using the weak label from other clustering method as a self- or unsupervised learning to improve the clustering performance. For example,
\cite{S2C} use the label of spectral clustering as a semi-supervised label to guide the convolutional subspace clustering network. 
However, these methods will be destroyed if the guided clustering performs poorly. 
Inspired by the success of using pairwise constraints in semi-supervised classification and the higher value of BCA, we intend to utilize the weak label of pairwise connection from the unsupervised clustering as a bridge to combine two clustering methods and promote the clustering performance. 

In the paper, we propose a bilevel clustering optimization framework with a learnable soft mask to combine two different clustering methods and give a specific case based on subspace clustering.
More precisely, we make the following contributions:\\
(1) we first propose a general Bilevel Clustering Optimization with a learnable soft mask for fusing different clustering optimizations.\\
(2) we introduce a General Masked Subspace Clustering method to combine different unsupervised or semi-supervised clustering results to subspace clustering via a soft mask.\\ 
(3) we provide a novel Recursive Masked Subspace Clustering method with a learnable soft mask to gradually and recursively enhance the different pairwise connections while keeping the property of self-expressiveness.\\
(4) our methods outperform other subspace clustering methods on several commonly used datasets.

\section{Related Work}
\subsection{Sparse Subspace Clustering}
Subspace clustering aims at grouping data points into different clusters, and each point is a linear combination of others in the same cluster. In order to character the property of self-expressiveness and sparsity, most existing methods apply low-rank or sparse-based regularizers on the affinity matrix to approximate a diagonal block structure. 

Given the data points as $X \in \mathcal{R}^{N\times d}$ with $N$ number of samples and $d$ dimension of features,
existing subspace clustering methods try to find an affinity matrix $C \in \mathcal{R}^{N\times N}$ via the following optimization~\cite{BDR}:
\begin{equation}
\min_{Z} f(Z,X) \quad s.t. \quad X = XZ, \quad Z\in \Omega.
\end{equation}
where $f(Z, X)$ denotes the regularization term that controls the structure and sparsity of the affinity matrix, $\Omega$ denotes the domain, which always comes to $\{Z|diag(Z)=\textbf{0} \}$. Common used regularizers in subspace clustering are low-rank and sparse-based priors, such as $\ell_1$-norm~\cite{SSC}, $\ell_0$-norm~\cite{SSC-O}, nuclear norm~\cite{LRR} or even more complex regularizations~\cite{BDR}. If the subspaces are independent, the affinity matrix obtained from these norms will satisfy the block diagonal property, which leads to a good clustering result. However, it is challenging to characterize the affinity matrix in complex data. Instead of designing novel mathematical structures, we focus on inducing a data-driven and learnable soft mask as pairwise constraints to make the affinity matrix tighter and sparser. 

\subsection{Bilevel Optimization}
Bilevel optimization\cite{BO} contains two subproblems: the outer optimization and the inner optimization, the solution of one is the input of the other. In the area of machine learning and computer vision, lots of models or algorithms belong to bilevel optimization, such as EM algorithm~\cite{EM}, adversarial learning~\cite{GANs}, meta-learning~\cite{meta}, neural architecture search, and reinforcement learning~\cite{BO}.
We focus on fusing two clustering methods and making them mutual promotion.

\section{A Unified Bilevel Clustering Optimization Framework}
\begin{figure}[t]
\centering
\includegraphics[width=8.5cm]{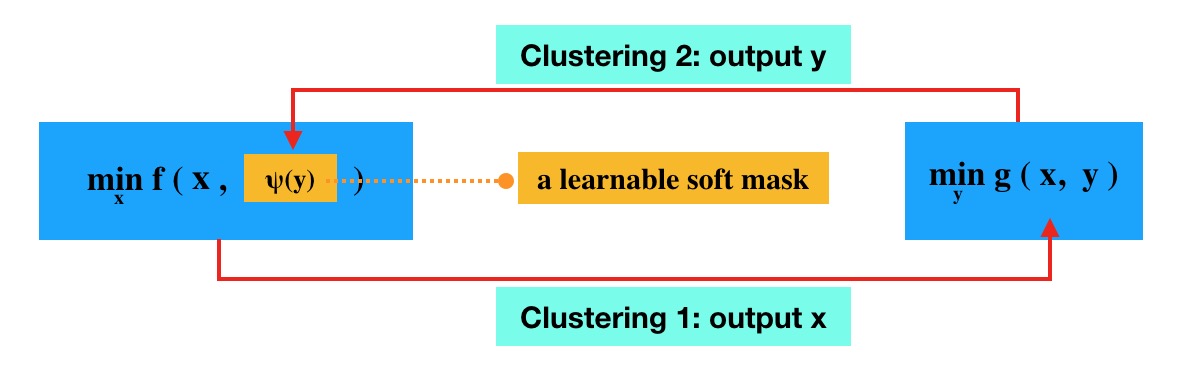}
\caption{A unified bilevel optimization framework for clustering with a learnable soft mask.}
\label{fig:com}
\end{figure}

To strengthen the connection between similar data points and weaken the connection between distant data points, we first provide a unified Bilevel Clustering Optimization (BCO) framework to combine two different clustering methods through a learnable soft mask as follows:
\begin{equation}
\begin{array}{ccc}
\min_{x} f(x, \psi(y))\\
s.t. \quad y = \arg\min_y g (x,y) \\
\end{array}
\label{eq:bco}
\end{equation}

where $f$ and $g$ represent two different clustering methods with the corresponding optimized parameters $x$ and $y$. $M^s = \psi(y)$ denotes the soft mask with the optimal $y$ from the second clustering optimization $g$. According to the clustering result from the second clustering method, if $i$ and $j$ belong to the same cluster, we assign a higher value of $M^s_{ij}$, otherwise with a lower value, such as
\begin{equation}
\left \{
\begin{array}{rcl}
M^s_{ij} \geqslant 1, & \mbox{if} & i \mbox{ and } j ~\mbox{belong to the same cluster}\\
M^s_{ij} < 1, & \mbox{if} & i \mbox{ and } j ~\mbox{belong to different clusters}.
\end{array}
\right .
\end{equation}
There exist many functions that maintain the property of the soft mask, we will give a simple example in our method.

\section{Masked Subspace Clustering Methods}
Before introducing the specific case of BCO, we first provide two preliminary subspace clustering methods with two types of prior masks. 
\subsection{Basic Masked Subspace Clustering}
It is well known that the original Sparse Space Clustering (SSC) minimizing the following problem:
\begin{equation}
\min_{Z} \|X-XZ \|^2_2 +\lambda \|Z\|_0 \quad s.t. \quad diag(Z) = \textbf{0}.
\label{eq:SSC}
\end{equation}
Different from adding a constraint on the diagonal elements in model~(\ref{eq:SSC}), we reformulate SSC as a hard mask on the entire matrix with a pairwise element product $\odot$, which leads to the Basic Masked Subspace Clustering (BMSC) model:
\begin{equation}
\min_{Z,C} \|X-XZ \|^2_2 + \lambda \|Z\|_0 \quad s.t. \quad Z = M^h \odot C.
\label{eq:bmsc}
\end{equation}
where $M^h$ is a matrix, whose diagonal elements are equal to 0, and others are equal to 1. We named $M^h$ as the hard matrix.
\subsection{General Masked Subspace Clustering}
Different from the hard mask in BMSC, we propose a soft mask to restrict the pairwise constraint. In $M^h$, if two different samples $x_i$ and $x_j$ belong to the same cluster, $M^h_{ij}=1$, otherwise $M^h_{ij}=0$. Instead of using such strong restrict, we use a soft mask in the following General Masked Subspace Clustering (GMSC) model to fuse soft pairwise constraints:
\begin{equation}
\begin{array}{cc}
\min_{C,Z} \frac 1 2 \|X-XZ\|_2^2 + \lambda \|Z\|_0 + \frac 1 2 \|M^s \odot C\|^2_2,\\
s.t. \quad Z = M^h \odot C.
\label{eq:gmsc}
\end{array}
\end{equation}
where $M^s$ is a soft mask, $M^h$ is the hard mask in model~(\ref{eq:bmsc}).\\
\textbf{Why use soft mask?} There are two reasons why should we choose a soft mask: 
\begin{enumerate}
\item Reducing error: if without truth labels, any hard mask from other clusterings may lead to errors.
\item Reducing noise and outliers: a hard mask may amplify the effect of outliers and noise.
\end{enumerate}
Thus, using a soft mask is beneficial to enhance the pairwise connection, avoid possible errors and reduce the sensitivity of noisy points and outliers. 

\textbf{Remark: } due to the existing of noise and outliers, we provide another version of GMSC:
{\small{
\begin{equation}
\begin{array}{cc}
\min_{C,Z,\varepsilon,O} \frac 1 2 \|X-XZ-\varepsilon-O\|_2^2 + \frac 1 2 \|M^s \odot C\|^2_2 + \lambda \|Z\|_0 + \lambda_1 \| \varepsilon \|^2_2 + \lambda_2 \|O\|_0\\
s.t.  \quad Z = M^h \odot C.
\label{eq:gmsc-n}
\end{array}
\end{equation}
}}
where $\varepsilon$ present the noise and $O$ present the outliers. We will use this version if the noise and outliers are not neglectable in the dataset.

\begin{figure}[ht]
    \centering
    \begin{minipage}[t]{0.49\textwidth} 
        \centering
        \begin{algorithm}[H]
            \caption{for BMSC/GMSC}
            \label{alg:gmsc}
            \begin{algorithmic}
               \State \textbf{Input:} $X$, $M^h$, $C^0, Q^0, U_1^0, U_2^0, k=0$.
               \State \textbf{Parameters:} $\lambda, \mu, p, T$.
               \State \textbf{Output:} $C^{k+1},Z^{k+1},Q^{k+1}$.
                \While{$k < T $}
                  \State update $Z^{k+1}$ with Eq.(\ref{Eq:b-Z}).
                    \State update $C^{k+1}$ with Eq.(\ref{Eq:b-C})/Eq.(\ref{Eq:C}).
                    \State update $Q^{k+1}$ with Eq.(\ref{Eq:b-Q}).
                    \State $U_1^{k+1} \leftarrow U_1^k + \mu(Z^{k+1}-M_h \odot C^{k+1})$.
                    \State $U_2^{k+1}  \leftarrow U_2^k + \mu(Q^{k+1}-Z^{k+1} )$.
                    \State $\mu \leftarrow \mu p$.
                    \If {converges}
                    \State break
                    \EndIf
                \State $k\leftarrow k+1$
                \EndWhile
            \end{algorithmic}
        \end{algorithm}
    \end{minipage}%
    \hfill
    \begin{minipage}[t]{0.495\textwidth} 
        \centering
        \begin{algorithm}[H]
            \caption{for RMSC}
            \label{alg:gmsc}
            \begin{algorithmic}
               \State \textbf{Input:} $X$, initial soft mask $M^s_{old},k=0.$
               \State \textbf{Parameters:} $P,\epsilon, T.$
               \State \textbf{Output:} $C,Z,Q.$
                \While{$k < T $}
                    \State $[C,Z,Q] = GMSC(X,M^h,M^s_{old},P)$
                    \State $\phi(Z)=\frac {|Z|+|Z|^T} 2$
                    \State $Q_{Lable} = spectral\_clustering(\phi(Z))$
                    \State $M^s_{new} = \psi(Q_{Lable})$
                    \If {$\|M^s_{new} - M^s_{old}\|_{\infty}<\epsilon$}
                    \State break
                    \Else
                    \State $M^s_{old} = M^s_{new}$
                    \EndIf
                \State $k\leftarrow k+1$
                \EndWhile
            \end{algorithmic}
        \end{algorithm}
    \end{minipage}
\end{figure}

\subsection{Recursive Masked Subspace Clustering}
The performance of GMSC is highly dependent on the soft mask derived from prior knowledge. To relieve the situation, we propose a Recursive Masked Subspace Clustering (RMSC) model to update the mask:
\begin{equation}
\begin{array}{ccc}
& \underset{Z\in \mathcal{Z}} {\min}  \frac 1 2 \|X-XZ\|_2^2 + \lambda \|Z\|_0 + \frac 1 2 \|\psi(Q) \odot Z\|^2_2 \\
s.t. & \quad F = \underset {F \in \mathcal{F}} {\arg\min} ~F^T\phi(Z)F\\
& Q = \underset {Q \in \mathcal{Q}} {\arg\min} \sum_{k=1}^{K} \sum_{i=1}^{N} \|F_i Q_{i,k} - m_k \|_2^2 
\label{eq:rmsc}
\end{array}
\end{equation}
where $F$ is the spectral embedding based on the adjacent matrix $\phi(Z)$, $Q$ is the K-means clustering on $F$. For details,
\begin{equation*}
\mathcal{Z} = \{Z|Z\in \mathcal{R}^{N\times N}, diag(Z)=\textbf{0} \}, \mathcal{F} = \{F|F\in \mathcal{R}^{N\times K} \}, \mathcal{Q} = \{Q|Q\in \{\textbf{0,1}\}^{N\times K} \},
\end{equation*}
\begin{equation}
m_k = \frac {\sum_{i=1}^N F_iQ_{i,k}} {\sum_{i=1}^N Q_{i,k}}, \psi(Q) = \beta Q^TQ + \frac 1 {\beta} (1-Q^TQ), \phi(Z) = \frac {|Z|+|Z|^T} 2.
\end{equation}
Let $x=Z, y=\{F,Q\}$, problem~(\ref{eq:rmsc}) is an example of bilevel clustering optimization~(\ref{eq:bco}).
It is clear that RMSC contains three subproblems, and each subproblem is equal to a mentioned or known problem:
\begin{enumerate}
\item Subproblem-$Z$: fix $Q$, update $Z$ $\rightarrow$ GMSC.
\item Subproblem-$F$: fix $Z$, update $F$ $\rightarrow$ Spectral embedding.
\item Subproblem-$Q$: fix $F$, update $Q$ $\rightarrow$ K-means clustering.
\end{enumerate}
Spectral clustering contains subproblem-$F$ and subproblem-$Q$, and RMSC is a way to combine subspace clustering and spectral clustering. Additionally, there is a strong relationship between BMSC, GMSC, and RMSC. If $M^s=\textbf{0}$, GMSC is degraded to BMSC; if the iteration time in RMSC equals 1, RMSC is equivalent to GMSC. Therefore, BMSC and GMSC are two simple cases of RMSC. 


\subsection{Algorithms for solving RMSC}
We use an alternative method to solve RMSC. As shown in Algorithm 2, with the initial soft mask, we first solve GMSC for the affinity matrix, then apply spectral clustering to obtain the clustering label. The solution for GMSC and BMSC are similar and will be presented in detail. \\
\textbf{Solution for BMSC:} we use an auxiliary variable $Q$ to reformulate problem ~(\ref{eq:bmsc}) as:
\begin{equation}
\begin{array}{ccc}
\min_{C,Z,Q} \frac 1 2 \|X-XZ\|_2^2 + \lambda \|Q\|_0\\
s.t. \quad Z = M^h \odot C, \quad Q = Z.
\end{array}
\end{equation}
Then we define the augmented Lagrangian function as follows:
\begin{equation}
\begin{array}{lll}
J &= \frac 1 2 \|X-XZ\|^2_2 + \lambda \|Q\|_0 + \langle U_1, Z-M^h\odot C \rangle\\
& + \frac {\mu} 2 \| Z-M^h \odot C \|^2_2 +  \langle U_2, Q-Z \rangle + \frac {\mu} 2 \|Q-Z\|^2_2.
\label{eq:b-J}
\end{array}
\end{equation}

\begin{itemize}
\item [*] \textbf{$Z-subproblem:$} let $\frac {\partial J} {\partial Z} = 0$, we have
\begin{equation}
Z = (X^TX+2\mu I)^{-1}(X^TX-U_1+\mu M^h\odot C+U_2+\mu Q).
\label{Eq:b-Z}
\end{equation}
where $I$ is a unit matrix. 
\item [*] \textbf{$C-subproblem:$} let $\frac {\partial J} {\partial C} = 0$, we have
\begin{equation}
C = (U/\mu + Z)\odot M^h.
\label{Eq:b-C}
\end{equation}
\item [*] \textbf{$Q-subproblem:$} according to the iterative hard thresholding algorithm for solving $\ell_0$-normed problem~\cite{L0-ht1,L0-ht2}, we have 
\begin{equation}
Q = \mathcal{H}_{\tau}(Z-U_2/\mu),
 \label{Eq:b-Q}
\end{equation}
where $\tau = \sqrt{2\lambda/\mu}$, $\mathcal{H}$ is a hard thresholding operator
\begin{equation*}
\mathcal{H}_{\tau}(t)= \left\{\begin{array}{c}t, \quad |t|> \tau, \\0, \quad |t|\leq \tau. \end{array}\right. 
\end{equation*}
\end{itemize}
\textbf{Solution for GMSC:} similar to BMSC, we rewrite problem~(\ref{eq:gmsc}) as:
\begin{equation}
\begin{array}{ccc}
\min_{C,Z,Q} \frac 1 2 \|X-XZ\|_2^2 + \lambda \|Q\|_0 + \frac 1 2 \|M^s \odot C\|^2_2\\
s.t. \quad Z = M^h \odot C, \quad Q = Z.
\end{array}
\end{equation}
And then we define the augmented Lagrangian function:
\begin{equation}
\begin{array}{lll}
J& =  \frac 1 2 \|X-XZ\|_2^2 + \lambda \|Q\|_0 + \frac 1 2 \|M^s \odot C\|^2_2+ \langle U_1, Z-M^h \odot C \rangle \\
& + \frac {\mu} 2 \| Z-M^h \odot C \|^2_2  + \langle U_2, Q-Z \rangle + \frac {\mu} 2 \| Q-Z \|^2_2. \\
\end{array}
\end{equation}
Subproblem-$Z$ and subproblem-$Q$ have the same solution in Equation~(\ref{eq:b-J}). For subproblem-$C$, let $\frac {\partial J} {\partial C} = 0$, we have
\begin{equation}
(Ms+\mu M^h) \odot C = (U_1+\mu Z)\odot M^h.
 \label{Eq:C}
\end{equation}
Hence, Algorithms 1 shows the detail for solving BMSC and GMSC, and the difference is the equation of solving the subproblem-$C$. We use 
\begin{equation}
\max \{|C^{k+1}-C^k|,|Z^{k+1}-Z^k|,|Q^{k+1}-Q^k|\}<10^{-6}
\end{equation}
 as the termination criterion of GMSC.
Additionally, the method for solving version~(\ref{eq:gmsc-n}) is the same as GMSC, with two more variables, which we do not repeat. 

\textbf{Theoretical convergence analysis:}
unfortunately, there is no global convergence analysis for the whole algorithm of solving RMSC, but each subproblem has a convergence analysis.
For subproblem-$Z$, GMSC is similar to the $\ell_0+\ell_2$-norm-based problem~\cite{L0_L2}. According to the theoretical proof, if $X^TX + M_s \odot M_s \succ 0$, the alternative algorithm for solving GMSC converges to a local minimum. Subproblem-$F$ has a global minimum if $\phi(Z)$ is positive semidefinite.  
For subproblem-$Q$, K-means convergents under some assumptions, as shown in \cite{kmeans-c1,kmeans-c2}.
\section{Experiments}
\subsection{Data description and experimental setting}
\begin{description}
    \item \textbf{Datasets:} We conduct experiments on several commonly used datasets, which include MNIST~\cite{MNIST}, USPS~\cite{USPS}, ORL~\cite{ORL}, COIL20, and COIL100~\cite{COIL}.
    \item \textbf{Baselines:} we first choose four low-rank and sparse-based subspace clustering methods LRR~\cite{LRR}, SSC~\cite{SSC}, SSC-OMP~\cite{SSC-O}, BDR~\cite{BDR}as the first baselines to show the effectiveness of our model that applies pairwise constraint to reshape the affinity matrix. Subsequently, we use three additional subspace methods EDSC~\cite{EDSC}, AE+EDSC, and Soft $S^3C$~\cite{S3C} to show the effectiveness of our model on different datasets.
    \item \textbf{Metrics:} we use the clustering accuracy and Normalized Mutual Information (NMI) to test the clustering performance.
    \item \textbf{Parameter setting:} there are several key parameters in our methods, different parameters lead to different effects, where
\begin{itemize}
\item $\lambda$:  determine the sparsity of affinity matrix.
\item $\beta$: balance the effect of soft mask.
\item $p$: related to the Lagrangian multiplier, local minimum and stability.
\end{itemize}
In our experiments, we set $\lambda \in \{10^{-3}, 5\times10^{-2}, 3\times10^{-4}, 10^{-3}, 10^{-4}\}$ on MNIST, USPS, ORL, COIL20, COIL100 respectively; 
$\beta$ is related to the reliability of the first clustering, we set $\beta=1.05$ for high confidence and $\beta=1.1$ for low confidence;
$p\in \{1.1,2\}$, we choose the best value of $p$. 
Appropriate parameter setting is a benefit to the clustering performance.
\end{description}

\subsection{Compare with the low-rank and sparse-based methods}
We randomly choose 1000 samples from the total MNIST dataset to test the performance. After 20 trials, we calculate the mean value and the standard deviation of the clustering accuracy, NMI, and the time for computing the affinity matrix in each method. Table~\ref{table:acc} shows that BMSC works with the least time and obtains higher accuracy than LRR, SSC, SSC-OMP, and BDR-B; GMSC has a higher clustering accuracy than BMSC but needs more time. RMSC achieves the highest values of accuracy and NMI, but with a moderate time cost. When compared with the mentioned low-rank and sparse-based models, RMSC has a significant improvement on clustering accuracy.

\setlength{\tabcolsep}{4pt}
\begin{table}[t]
\begin{center}
\caption{Clustering performance on MNIST.}
\label{table:acc}
\begin{tabular}{lccc}
\hline\noalign{\smallskip}
Models & Accuracy (\%) & NMI (\%) & Time (s) \\
\noalign{\smallskip}
\hline
\noalign{\smallskip}
LRR & 56.38 $\pm$ 3.26 & 53.64 $\pm$ 2.16  & 59.03 $\pm$ 5.27\\ 
SSC & 51.57 $\pm$ 3.55 & 53.82 $\pm$ 2.62 & 8.27 $\pm$ 1.63 \\
SSC-OMP & 51.48 $\pm$ 4.17 & 47.09 $\pm$ 2.83 & 17.79 $\pm$ 1.09 \\
BDR-B & 57.77 $\pm$ 4.45 & 61.41 $\pm$ 3.17 & 89.46 $\pm$ 10.35\\
\hline
BMSC & 58.97 $\pm$ 3.22 & 63.45 $\pm$ 2.18 & \textbf{0.88 $\pm$ 0.16}\\
GMSC & 59.38 $\pm$ 4.06 & 64.69 $\pm$ 2.64 & 5.61 $\pm$ 0.65 \\
RMSC & \textbf{60.32 $\pm$ 4.12} & \textbf{64.94 $\pm$ 3.10} & 23.10 $\pm$ 11.12\\
\hline
\end{tabular}
\end{center}
\end{table}
\setlength{\tabcolsep}{1.4pt}
\vspace{-2em}
\setlength{\tabcolsep}{4pt}
\begin{table}[t]
\begin{center}
\caption{Clustering accuracy (\%) on different methods and datasets.}
\label{table:errs}
\begin{tabular}{lcccc}
\hline\noalign{\smallskip}
Models & USPS & ORL & COIL20 & COIL100 \\
\noalign{\smallskip}
\hline
\noalign{\smallskip}
LRR & 77.40 & 66.50 & 69.79 & 46.82 \\
SSC & 56.90 & 70.50 & 85.17 & 55.10 \\
SSC-OMP & 64.60 & 62.95 & 70.14 & 32.71 \\
EDSC & - & 72.75 & 85.14 & 61.83\\
AE+EDSC & - & 73.75 & 85.21 & 61.12 \\
Soft $S^3C$ & - & 74.00 & 88.13 & 58.29 \\
BDR & 61.30 & 61.25 & 84.03 & - \\
\hline
BMSC & 65.60 & 66.00 & 80.49 & 56.00 \\
GMSC & \underline{77.60} & 74.25 & 82.78 & 55.65 \\
RMSC-v1 & 76.9 & \underline{83.00} & \textbf{91.94} & \textbf{65.58} \\
RMSC-v2 & \textbf{78.30} & \textbf{86.00} & \underline{91.67} & \underline{64.04} \\
\hline
\end{tabular}
\end{center}
\end{table}
\setlength{\tabcolsep}{1.4pt}

\subsection{Comparisons on different methods and datasets.}
To further validate the effectiveness of our methods, we compare against several models on multiple datasets, such as USPS, ORL, COIL20, and COIL100. Here, to distinguish the different versions of GMSC in RMSC, we denote RMSC-v1 as a usage of model~(\ref{eq:gmsc}), and RMSC-v2 as the version of model~(\ref{eq:gmsc-n}). Table~\ref{table:errs} demonstrates that GMSC has greater accuracy than BMSC, and RMSC obtains the highest clustering accuracy among all listed datasets. For datasets USPS and ORL, after taking the consideration of noise and outliers, RMSC-v2 obtains an improvement in the clustering accuracy. 
However, for COIL20 and COIL100, the clustering accuracy drops a little. If the noise and outlier are minimal, we suggest using RMSC-v1.  Moreover, to better understand the effectiveness of RMSC, we will give an insight and comparison between BMSC, GMSC, and RMSC.

\begin{figure}[ht]
    \centering
    \begin{minipage}[t]{0.45\textwidth} 
        \includegraphics[width=\textwidth]{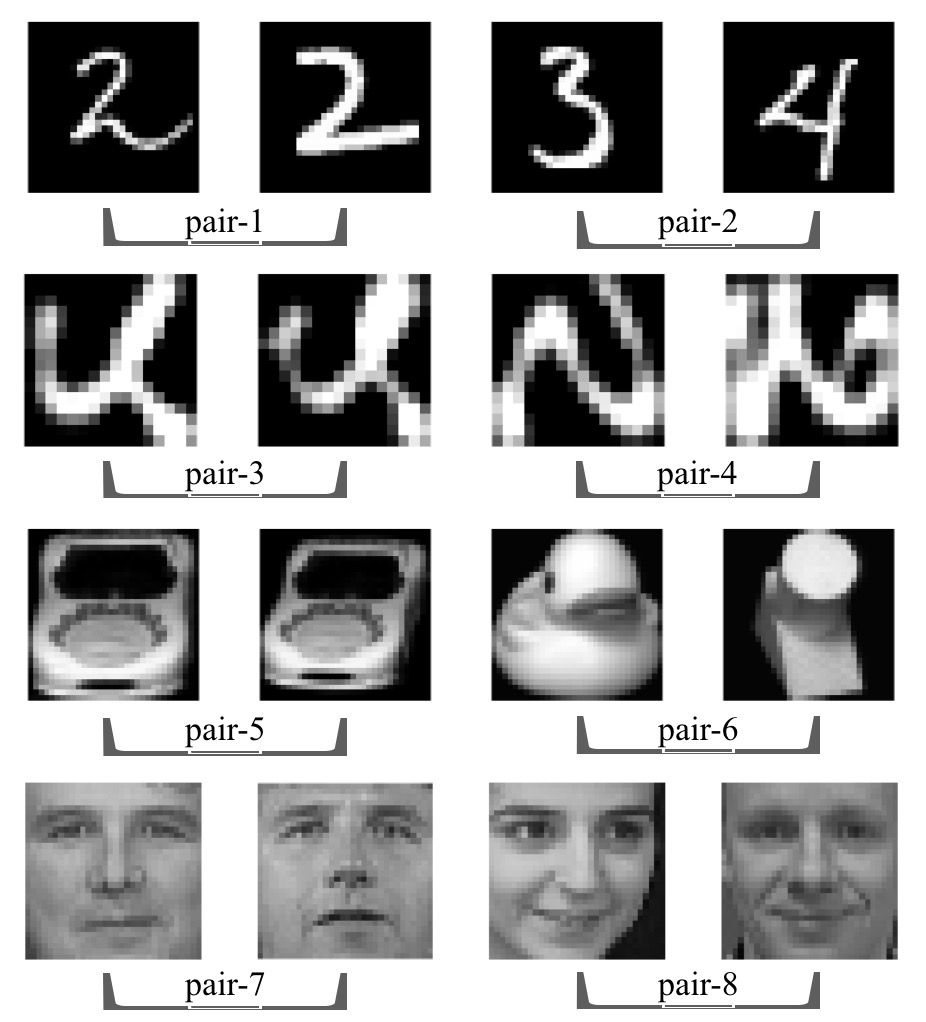} 
        \caption{Pairs visualization. Eight pairs of images from MNIST, USPS, ORL, and COIL20 were selected to evaluate the values of the affinity matrices. For each dataset, there have two pairs: one with the same class and the other with different classes.}
        \label{fig:pairs}
    \end{minipage}%
    \hfill
    \begin{minipage}[t]{0.48\textwidth} 
        \centering
        \raisebox{3cm}{
        \begin{tabular}{ l c c c c } 
            \hline\noalign{\smallskip}
            Types & Pairs & BMSC & GMSC & RMSC \\
            \noalign{\smallskip}
            \hline
            \multirow{3}{4em}{Same Class} & pair-1 & 0.0001 & 0.0399 &  \textbf{0.0472}\\ 
            & pair-3  & 0.0385 & 0.0827 &  \textbf{0.1102} \\ 
            & pair-5  & 0.0332 & 0.0576 &  \textbf{0.0879} \\ 
            & pair-7 & 0.0000 & 0.0001 &  \textbf{0.0204} \\
            \hline
            \multirow{3}{4em}{Different Class} & pair-2 & 0.0766 & 0.0421 & \textbf{0.0000} \\ 
            & pair-4  & 0.0399 & 0.0391 &  \textbf{0.0001} \\ 
            & pair-6  & 0.0736 & 0.0397 &  \textbf{0.0003} \\ 
            & pair-8 & 0.0389 & 0.0270 & \textbf{0.0002}\\
            \hline
        \end{tabular}
        }
        \captionof{table}{The values of the affinity matrices on different pairs and models: a larger value means a closer relation. GMSC and RMSC help to strengthen the connection between the same class and weaken the connection between different classes.}
        \label{tab:example}
        \vfill
    \end{minipage}
\end{figure}

\begin{figure*}[ht]
\centering
\includegraphics[width=10cm,height=5cm]{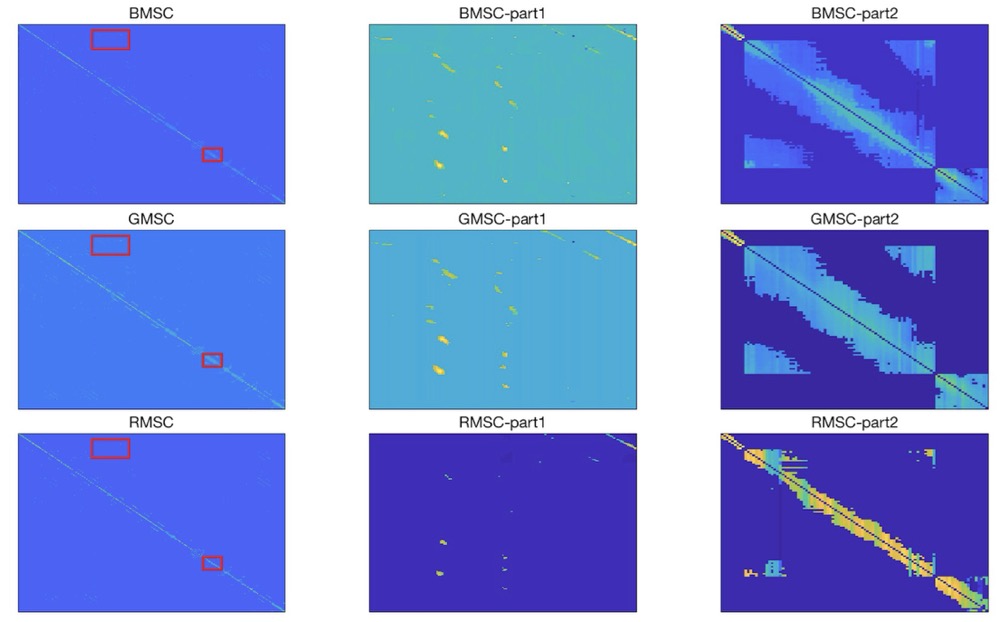}
\caption{Visualization of the affinity matrix from BMSC, GMSC, and RMSC: RMSC-parts have the highest sparsity and the strongest connections.}
\label{fig:com}
\end{figure*}

\subsection{Comparisons between BMSC, GMSC and RMSC}
Table 1 shows that RMSC outperforms GMSC and GMSC performs better than BMSC. To evaluate the effect of the soft mask and the recursive scheme on the pairwise connection, we demonstrate the visualization of the affinity matrix in Fig.\ref{fig:com}. We compared the affinity matrices obtained from BMSC, GMSC, and RMSC on data COIL20. The zoom-in areas show that the affinity matrices derived from GMSC have a higher sparsity than BMSC, which means the prior mask contributed to strengthening the connection. The affinity matrices obtained from RMSC have the highest sparsity overall and the closest relation on tight pairs, which means the recursive scheme works for strengthening the relationship between similar pairs and weakening the connection between distant pairs. 

More precisely, we compare the values of the affinity matrices and present eight pairs of images (Fig.\ref{fig:pairs}) to show the effect of GMSC and RMSC. A larger value means a closer relation. Table 3 shows, these pairs with the same class on BMSC have the smallest values, while RMSC obtains the highest values. The images of pair-1 belong to the same class, they suppose to have a strong relationship, but in BMSC, the value is lower to 0.0001. Fortunately, GMSC and RMSC boost the value. 
On the contrary, the pairs with different classes on BMSC have the highest values, RMSC achieves the lowest values, and GMSC is in the middle. Images in pair-2 and pair-6 are different and have large values on BMSC, but GMSC and RMSC help to reduce the values.
In a word, GMSC and RMSC can strengthen the connection between the same class, and weaken the connection between different classes. It proves that the soft mask and the recursive scheme of updating the mask play a positive role, and regarding the pairwise connection from unsupervised clustering as a weak constraint is valuable.
\vspace{-1em}
\section{Conclusion}
\vspace{-1em}
We propose a general bilevel clustering optimization that utilizes a learnable soft mask to combine two clustering methods and promote the clustering performance. A recursive masked subspace clustering which integrates sparse subspace clustering and spectral clustering was presented as a specific case. On the aspect of subspace clustering, we introduce three types of masks: a hard mask, a soft mask, and a learnable soft mask. Based on these marks, we build three masked subspace clustering models: BMSC, GMSC, and RMSC, and provide the corresponding algorithms. Numerical experiments evaluate the effectiveness of our models. Our methods obtain a significant improvement compared to the low-rank and sparsity-based subspace clustering models; the soft mark and the recursive scheme for updating the affinity matrix help to strengthen the connection between the same class and weaken the relation between different categories.

\vspace{-1em}
\bibliography{RMSC}
\vspace{-1em}
\bibliographystyle{splncs04}
\end{document}